\documentclass[pdflatex, sn-mathphys-num]{sn-jnl}

\usepackage{threeparttable,graphicx} 
\usepackage{graphics}
\usepackage{multirow}%
\usepackage{amsmath,amssymb,amsfonts}%
\usepackage{amsthm}%
\usepackage{mathrsfs}%
\usepackage[title]{appendix}%
\usepackage{xcolor}%
\usepackage{textcomp}%
\usepackage{manyfoot}%
\usepackage{booktabs}%
\usepackage{algorithm}%
\usepackage{algorithmicx}%
\usepackage{algpseudocode}%
\usepackage{listings}%
\usepackage{tcolorbox}%
\usepackage{mdframed}%

\newcommand{\wstd}[2]{#1\,{\tiny$\pm #2$}}           
    
\newcommand{\mch}[1]{\begin{tabular}{@{}c@{}}#1\end{tabular}}
\newcommand{\wstdg}[2]{#1\,{\footnotesize$#2$}}
\newcommand{\bwstdg}[2]{\textbf{#1}\,{\footnotesize$#2$}}

\newtcolorbox{templatebox}[1][]{
    colback=orange!10,
    colframe=orange!50,
    boxrule=0.5pt,
    arc=2pt,
    left=6pt,
    right=6pt,
    top=6pt,
    bottom=6pt,
    fontupper=\footnotesize,
    #1
}

\theoremstyle{thmstyleone}%
\theoremstyle{thmstyletwo}%

\theoremstyle{thmstylethree}%

\begin{document}

\title[Holistic Artificial Intelligence in Medicine; improved performance and explainability]{Holistic AI in Medicine; improved performance and explainability}

\author[1]{\fnm{Periklis} \sur{Petridis}}\email{periklis@mit.edu}

\author[1]{\fnm{Georgios} \sur{Margaritis}}\email{geomar@mit.edu}

\author[1]{\fnm{Vasiliki} \sur{Stoumpou}}\email{vasstou@mit.edu}

\author*[2]{\fnm{Dimitris} \sur{Bertsimas}}\email{dbertsim@mit.edu}

\affil*[1]{\orgdiv{Operations Research Center}, \orgname{Massachusetts Institute of Technology}, \orgaddress{\city{Cambridge}, \state{MA}, \country{USA}}}

\affil[2]{\orgdiv{Sloan School of Management}, \orgname{Massachusetts Institute of Technology}, \orgaddress{\city{Cambridge}, \state{MA}, \country{USA}}}

\abstract{With the increasing interest in deploying Artificial Intelligence in medicine, we previously introduced HAIM (Holistic AI in Medicine), a framework that fuses multimodal data to solve downstream clinical tasks. However, HAIM uses data in a task-agnostic manner and lacks explainability. To address these limitations, we introduce xHAIM (Explainable HAIM), a novel framework leveraging Generative AI to enhance both prediction and explainability through four structured steps: (1) automatically identifying task-relevant patient data across modalities, (2) generating comprehensive patient summaries, (3) using these summaries for improved predictive modeling, and (4) providing clinical explanations by linking predictions to patient-specific medical knowledge. Evaluated on the HAIM-MIMIC-MM dataset, xHAIM improves average AUC from 79.9\% to 90.3\% across chest pathology and operative tasks. Importantly, xHAIM transforms AI from a black-box predictor into an explainable decision support system, enabling clinicians to interactively trace predictions back to relevant patient data, bridging AI advancements with clinical utility.}


\maketitle

\section{Introduction}\label{sec1}

Machine learning (ML) has grown rapidly over the last two decades. This growth has shown substantial promise for its application in critical real-world domains, including clinical settings and healthcare operations. Although the adoption of machine learning techniques in safety-critical applications initially faced inertia and skepticism, recent progress in model safety, fairness, and interpretability \cite{amann_explainability_2020,holzinger_information_2022,mehrabi_survey_2021}, as well as improvements in performance \cite{singhal_large_2023,singhal_toward_2025,saab_capabilities_2024}, has generated significant interest among clinicians and medical experts \cite{thirunavukarasu_large_2023,wang_perspective_2024}. Now, with the advent of foundational models (FMs) and the demonstration of accessible, general-purpose capabilities from recent large language models (LLMs) \cite{bubeck2023sparks,brown2020language}, interest from practitioners has grown considerably \cite{poon_adoption_2025,american_medical_association_augmented_2024}. However, important challenges still limit the widespread adoption of such models. For example, \cite{bedi_testing_2025} reports that only 5\% of healthcare organizations have successfully deployed AI solutions in clinical practice, revealing a significant gap between the promise of research and real-world implementation.

Two of the key factors limiting the deployment of machine learning models in medical settings are inadequate predictive performance \cite{kelly2019key} and lack of explainability \citep{wang_perspective_2024,freyer_ethical_2024,abgrall_should_2024}, both of which reduce the credibility of these models among practitioners.

When it comes to model performance, a critical challenge is the need for domain adaptation of general-purpose pre-trained models in clinical settings, as clinical data is rarely available publicly, with the exception of efforts such as MIMIC \cite{soenksen_code_2022,johnson_mimic-iii_2016,johnson_mimic-iv_2023}. Relying solely on data from a single institution does not enable training of large models on sufficiently diverse and comprehensive datasets, thereby limiting their performance compared to other fields with public data availability. One of the approaches that have emerged is Federated Learning (FL) \citep{mcmahan2017communication}, which enables training on decentralized data from multiple organizations without violating data-sharing regulations, allowing access to more diverse data while keeping patient records local. However, FL systems add architectural and operational complexity \cite{teo2024federated}, and their results are often affected by the different cohorts' distributions, which degrades their performance \cite{rieke2020future}. Due to the above considerations, domain adaptation for the clinical domain remains a significant challenge.

When it comes to explainability, widely used clinical models such as ClinicalBERT \cite{huang2020clinicalbertmodelingclinicalnotes} often operate as “black boxes.” While these models are effective for classification tasks, they provide little insight into their decision-making processes \cite{zhang_uncovering_2024}. This opacity poses a barrier to clinical implementation, where understanding the rationale behind AI-generated recommendations is critical for responsible care delivery. Current evaluation approaches still emphasize accuracy (used in 95.4\% of studies), while overlooking clinical utility, interpretability, and deployment considerations \cite{bedi_testing_2025}. Furthermore, traditional explainability techniques in medical AI, such as feature attribution and saliency maps \cite{amann_explainability_2020,holzinger_information_2022}, often fail to yield clinically meaningful narratives.

The goal of this work is to provide a framework for predicting pathologies and clinical outcomes that addresses these two central limitations: predictive performance and explainability. Existing methodologies typically excel at one objective while compromising the other. One notable example is HAIM (Holistic Artificial Intelligence in Medicine) \cite{soenksen_integrated_2022}, a state-of-the-art discriminative framework for multimodal clinical prediction that integrates data from various sources and has demonstrated high performance across multiple tasks, but operates as a "black box". More recently, Generative AI has been used to improve interpretability in clinical predictions \cite{xie2024me, savage2024diagnostic}, but these approaches have been shown to perform worse than traditional, albeit black-box, ML models \citep{chen_clinicalbench_2024, brown_not_2024}. This presents a critical gap in healthcare AI: the need for systems that achieve both high predictive performance and clinical interpretability.

HAIM, for instance, demonstrates strong predictive performance but lacks interpretability and faces practical challenges with data volume. The framework combines clinical notes, tabular data, time-series measurements, and medical images into a unified architecture, but processes entire patient histories indiscriminately, including extensive records from multiple visits, tests, and notes that may be irrelevant to the specific prediction task. This leads to inefficiencies, as clinical notes must be truncated to fit fixed input lengths (e.g., ClinicalBERT's token limits), and when records exceed these constraints, embeddings are averaged or concatenated, introducing noise and diluting important clinical signals.

On the other hand, recent work leveraging Generative AI for interpretability has primarily explored directions
such as clinical question answering \cite{chen_clinicalbench_2024}, document summarization for administrative efficiency \cite{kim_medical_2025}, clinical trial matching \cite{liu_lost_2023}, and reducing hallucinations via retrieval-augmented generation \cite{min_factscore_2023,xiong_benchmarking_2024}. Despite the rapid progress of generative LLMs—including models tailored for medicine—multiple studies have shown that the best performance on clinical classification tasks is still achieved by fine-tuning traditional ML models or discriminative LLMs (e.g., XGBoost, ClinicalBERT) on private hospital data \cite{brown_not_2024,chen_clinicalbench_2024,takita_systematic_2025,wornow_ehrshot_2023,fleming_medalign_2023}. As \cite{chang_survey_2024} observes, most current applications of LLMs in medicine focus on question answering and standardized exams (e.g., USMLE), rather than clinical prediction, where discriminative models still excel. Generative models also continue to hallucinate plausible but incorrect medical content \cite{bang2023multitask,kim_medical_2025}, creating major safety concerns in healthcare settings \cite{kim_medical_2025,miura_improving_2021}.

\begin{figure*}[t]
    \centering
    \includegraphics[width=\textwidth]{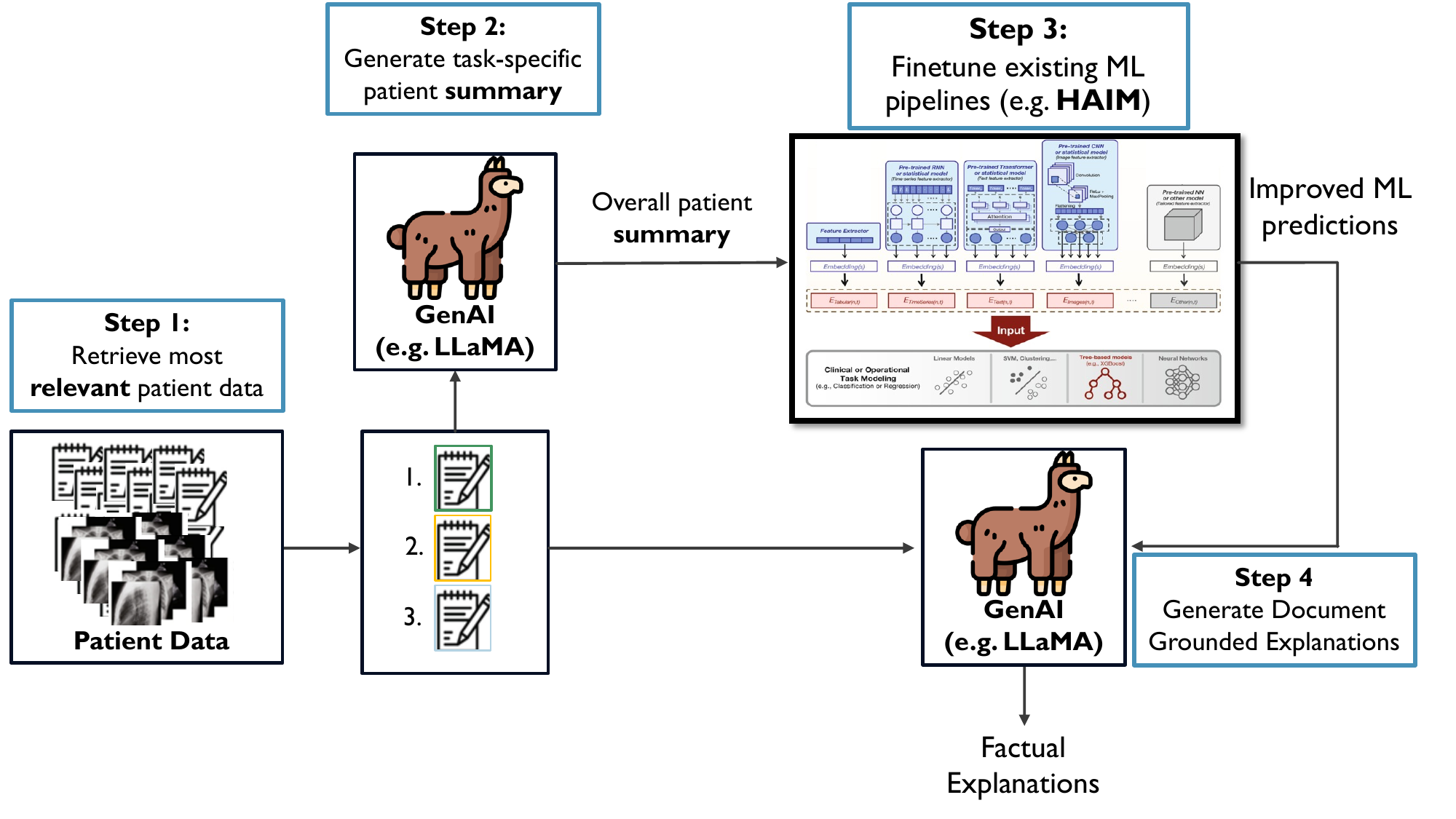}
    \caption{The xHAIM pipeline. (1) Task definition and relevant patient data identification using semantic similarity, (2) creation of comprehensive task-specific summaries, (3) enhanced HAIM predictions using focused input data, and (4) generation of explainable insights by combining predictions with relevant medical knowledge. This approach addresses challenges in data volume management, model interpretability, and medical knowledge integration.}\label{fig:pipeline}
\end{figure*}

To address the limitations of both paradigms, we propose a hybrid framework that aims to deliver the best of both worlds by improving predictive performance and explainability—two pillars essential for successful clinical deployment. We present xHAIM (Explainable Holistic AI in Medicine), an extension of the HAIM framework that leverages generative AI to enhance discriminative models using a four-step process (illustrated in Figure~\ref{fig:pipeline}). This process involves: \textbf{a)} automatically identifying task-relevant patient data across modalities using semantic similarity, \textbf{b)} generating focused clinical summaries that preserve essential information while filtering noise, \textbf{c)} improving predictive performance by using these curated summaries rather than potentially noisy embeddings, and \textbf{d)} providing clinically grounded explanations by augmenting predictions with relevant medical knowledge.

By using LLMs for intelligent preprocessing and post-hoc explanation generation, xHAIM addresses the interpretability limitations of discriminative models while preserving their superior predictive power. Rather than replacing existing clinical ML systems, our framework serves as a natural extension that enhances model inputs, regardless of data modality or structure. In contrast to recent LLM-based systems that use generative models for end-to-end tasks, xHAIM utilizes GenAI auxiliarily—leveraging its strengths for summarization and explanation while relying on lightweight, fine-tuned discriminative models for downstream prediction \cite{soenksen_integrated_2022}.

In the following sections, we present experimental evidence demonstrating xHAIM’s improvements in predictive performance and interpretability, followed by a detailed methodology and a discussion of its clinical implications.

\section{Results}\label{sec2}

In this section, we present experimental results in two parts: first comparing xHAIM's predictive performance against HAIM as a baseline, then evaluating its explainability through both manual annotation and LLM-as-a-judge automation.

\subsection*{Dataset and Clinical Tasks}\label{subsec:setup}

Our experimental evaluation is based on the HAIM-MIMIC-MM dataset \cite{soenksen_code_2022}, a comprehensive multimodal clinical dataset derived from the MIMIC-IV critical care database. The dataset encompasses information about patients in the Intensive Care Unit (ICU) from four distinct data modalities: \textbf{a)} clinical notes containing detailed patient narratives and assessments from Radiology, Echocardiogram and EKG reports, \textbf{b)} tabular data primarily including demographics, \textbf{c)} time-series monitoring data capturing continuous physiological measurements, such as vital signs and laboratory values, and \textbf{d)} medical images primarily consisting of chest X-rays and radiological findings. This multimodal collection comprises 34,537 samples spanning 7,279 unique hospitalizations across 6,485 patients, providing a robust foundation for evaluating clinical prediction tasks.

For these experiments, we leverage the full available cohort. Specifically, for each task of interest, we retain the subset of patients for whom the task-specific ground-truth label is available. Each ICU stay is treated as a separate data point: we randomly select one scan from the stay and use its timestamp as the decision time point. All available medical information (tabular data, time series, images and notes) prior to that time point is then used in the training process of the predictive model for the downstream task.

The data is split in 80/20 train-test set and each experiment is run for $5$ different splits to confirm the statistical validity of our results. Through our experiments, we also ensure that each patient's entries can only belong in one of the training and test sets. 

We focus our evaluation on five critical clinical prediction tasks that represent diverse aspects of patient care and prognosis in intensive care settings: pleural effusion detection, cardiomegaly classification, pneumonia diagnosis, 48-hour mortality prediction, and length of stay prediction. These tasks are widely employed benchmarks for AI performance in clinical practice \cite{chen_toward_2023,wornow_ehrshot_2023}, encompassing both diagnostic challenges (e.g., identifying pathologies from multimodal data) and prognostic assessments (predicting patient outcomes and resource utilization). The number of data points retained per task is displayed in Table \ref{tab:num_patients_per_task}. 

The evaluation metrics we use include area under the receiver operating characteristic curve (AUC) for classification tasks, alongside evaluations of the explanations, covering factual accuracy, citation correctness, and overall quality, using both manual annotations and LLM-as-a-judge assessments to ensure real-world clinical utility.

\begin{table*}[t]
    \centering
    \caption{Number of data points retained for each task.}
    \label{tab:num_patients_per_task}
    \resizebox{\textwidth}{!}{%
    \begin{tabular}{@{}lcccc@{}}
    \toprule
    \textbf{Pleural Effusion} & \textbf{Cardiomegaly} &
    \textbf{Pneumonia} & \textbf{Mortality} & \textbf{LOS} \\
    \midrule
    8,926 & 6,971 & 3,512 & 16,888 & 16,888 \\
    \bottomrule
    \end{tabular}}%
\end{table*}

\subsection*{Quantitative Performance Improvement in Clinical Prediction}\label{subsec:quant}

The original HAIM framework achieves an average AUC of 79.9\% across five clinical prediction tasks, but suffers from fundamental limitations in processing lengthy clinical texts (detailed in Section~\ref{sec4}). In contrast, xHAIM addresses these limitations through intelligent data curation and summary generation, achieving 90.3\% average AUC, substantially outperforming the baseline.

As shown in Table~\ref{tab:auc_results}, xHAIM significantly outperforms HAIM across all tasks, with the largest gains in pathology detection that relies heavily on clinical narrative: pleural effusion (+13.5\%), cardiomegaly (+16.3\%), and pneumonia (+19.4\%). For complex outcomes like mortality and length of stay, which are inherently challenging even for clinicians, xHAIM still achieves statistically significant improvements of +2.7\% and +1.9\% respectively. These results demonstrate that generative AI preprocessing can enhance discriminative models without replacing their predictive mechanisms, validating our hybrid approach for clinical applications.

\begin{table*}[t]
    \centering
    \caption{Average ROC AUC performance comparison across models and clinical conditions. Table entries show percentage AUC and standard error across $5$ runs. xHAIM-Qwen and xHAIM-Llama refer to the xHAIM pipeline using the Qwen and Llama models respectively, and xHAIM-FT-Qwen, xHAIM-FT-Llama refer to the corresponding finetuned versions.}
    \label{tab:auc_results}
    \resizebox{\textwidth}{!}{%
    \begin{tabular}{@{}lcccccc@{}}
    \toprule
    \textbf{Model} & \textbf{Pleural Effusion} & \textbf{Cardiomegaly} &
    \textbf{Pneumonia} & \textbf{Mortality} & \textbf{LOS} & \textbf{Average}\\
    \midrule
HAIM Baseline &  \wstdg{84.8}{\pm0.5} &  \wstdg{81.1}{\pm0.2} &  \wstdg{76.3}{\pm0.4} & \wstdg{82.0}{\pm0.2} & \wstdg{75.5}{\pm0.4}  & 79.9\\
    xHAIM-Qwen &  \wstdg{88.5}{\pm0.3} &  \wstdg{84.4}{\pm0.6} & \wstdg{84.9}{\pm0.6} & \wstdg{83.9}{\pm0.7} & \wstdg{75.3}{\pm0.6} & 83.4\\
   xHAIM-Llama &  \wstdg{90.6}{\pm0.3} &  \wstdg{85.9}{\pm0.5} & \wstdg{84.7}{\pm0.7} & \wstdg{82.7}{\pm0.3} & \wstdg{74.6}{\pm0.4} & 83.7\\
 xHAIM-FT-Qwen & \wstdg{97.1}{\pm0.1} & \wstdg{96.0}{\pm0.3} & \wstdg{94.1}{\pm0.4} & \bwstdg{84.7}{\pm0.4} & \bwstdg{77.4}{\pm0.4} & 89.9\\
xHAIM-FT-Llama & \bwstdg{98.3}{\pm0.1} & \bwstdg{97.4}{\pm0.2} & \bwstdg{95.7}{\pm0.3} & \wstdg{83.3}{\pm0.6}& \wstdg{76.9}{\pm0.6} & \textbf{90.3}\\
    \bottomrule
    \end{tabular}}%
\end{table*}

\subsection*{Explainability Evaluation with LLM-as-a-Judge}\label{subsec:eval_quality}
Beyond improving predictive performance, xHAIM generates clinically meaningful explanations that make AI reasoning transparent to healthcare professionals by identifying the key factors driving each prediction.

To systematically evaluate explanation quality, we developed a two-stage LLM-as-a-Judge framework calibrated against human expert annotations. This approach assesses three critical dimensions: \textbf{(a)} citation accuracy—verifying that references to patient documents are accurate and complete, \textbf{(b)} factual correctness—ensuring no unsupported medical claims, and \textbf{(c)} overall quality—evaluating coherence, conciseness, and clinical utility. The first stage employs an adversarial ``Clinical Documentation Quality Analyst'' to comprehensively identify potential issues across all dimensions. The second stage uses an ``Expert Medical Evaluation Analyst'' that evaluates explanations using the critique as a checklist while applying independent judgment based on established 1-5 scale rubrics (detailed in Appendix~\ref{subsec:evaluation_framework}).

\begin{figure}[ht]
    \centering
    \includegraphics[width=\columnwidth]{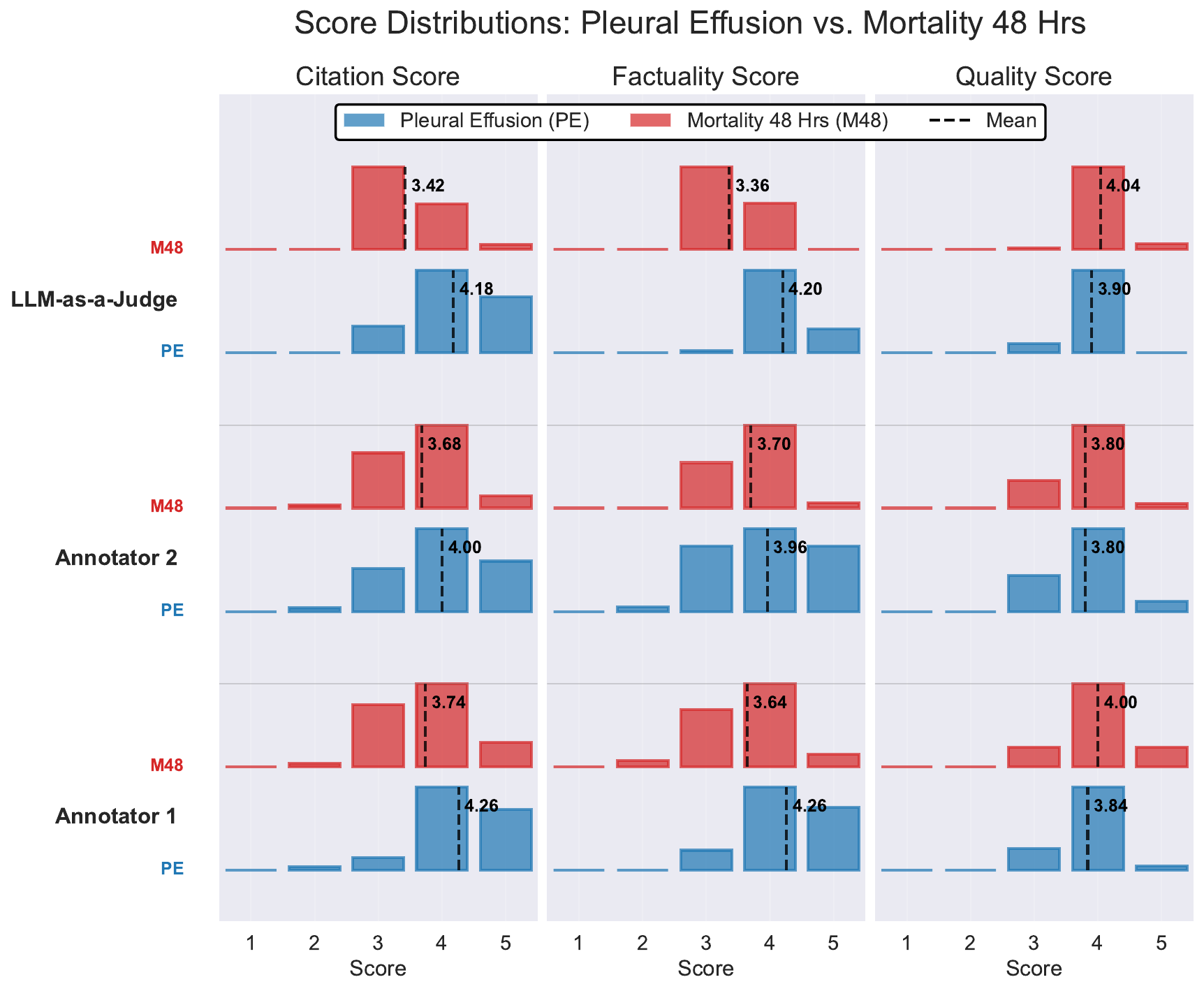}
    \caption{Distribution of explanation quality scores across evaluators and clinical conditions. Three-panel ridge plots show Citation (left), Factuality (middle), and Overall Quality (right) scores. Blue indicates Pleural Effusion, red indicates Mortality 48 Hrs. Red dashed lines show mean scores. The consistent lower scores for mortality prediction across all evaluators suggest this task presents greater explanation challenges.}
    \label{fig:eval_distributions}
\end{figure}

We validated this automated approach through systematic comparison with manual annotations on 50 explanations per task, demonstrating strong alignment with human judgment. As illustrated in Figure~\ref{fig:eval_distributions}, the framework provides evaluations comparable to human annotators across both diagnostic and operative tasks, though mortality prediction consistently receives lower scores, reflecting the inherent difficulty of explaining complex outcomes compared to specific pathologies like pleural effusion.

This validation enabled scaling to automatic evaluation of 1,000 explanations. Results (Tables~\ref{tab:eval_manual_detail} and~\ref{tab:eval_llm_summary}) show consistently high scores across all dimensions, with citation accuracy comparable to factuality.

The example explanations below demonstrate xHAIM's ability to generate concise, task-focused narratives that cite specific clinical findings rather than providing generic summaries.

%
%
\begin{table}[ht]
    \centering
    \caption{Detailed manual evaluation of explanation quality by human annotators (N=50).}
    \label{tab:eval_manual_detail}
    \begin{tabular}{llccc}
    \toprule
    & Evaluator & \mch{Citation \\ Score} & \mch{Factuality \\ Score} & \mch{Overall \\ Quality} \\
    \midrule
    \multirow{3}{*}{\rotatebox{90}{\footnotesize\mch{Pleural \\ Effusion}}}
    & Annotator 1 & \wstd{4.26}{0.69} & \wstd{4.26}{0.66} & \wstd{3.84}{0.47} \\
    & Annotator 2 & \wstd{4.00}{0.78} & \wstd{3.96}{0.83} & \wstd{3.80}{0.57} \\
    & LLM Judge & \wstd{4.18}{0.69} & \wstd{4.20}{0.45} & \wstd{3.90}{0.30} \\
    \midrule
    \multirow{3}{*}{\rotatebox{90}{\footnotesize \mch{Mortality \\ 48 Hrs}}}
    & Annotator 1 & \wstd{3.74}{0.72} & \wstd{3.64}{0.69} & \wstd{4.00}{0.57} \\
    & Annotator 2 & \wstd{3.68}{0.65} & \wstd{3.70}{0.54} & \wstd{3.80}{0.49} \\
    & LLM Judge & \wstd{3.42}{0.57} & \wstd{3.36}{0.48} & \wstd{4.04}{0.28} \\
    \\[-0.9ex]
    \bottomrule
    \end{tabular}
\end{table}

%
%
\begin{table}[ht]
    \centering
    \caption{LLM-as-a-Judge (ChatGPT) evaluation of explanation quality on the full test set (N=1000).}
    \label{tab:eval_llm_summary}
    \small
    \begin{tabular*}{\columnwidth}{@{\extracolsep{\fill}}lccc@{}}
    \toprule
    Condition & \mch{Citation \\ Score} & \mch{Factuality \\ Score} & \mch{Overall \\ Quality} \\
    \midrule
    Pleural Effusion & \wstd{4.15}{0.73} & \wstd{4.14}{0.68} & \wstd{3.85}{0.46} \\
    Mortality 48 Hrs & \wstd{3.54}{0.63} & \wstd{3.48}{0.67} & \wstd{3.86}{0.49} \\
    \bottomrule
    \end{tabular*}
\end{table}

\begin{templatebox}[title=Example 1: Explanation for Pleural Effusion]
\label{ex:pleural_explanation}
\footnotesize
\textit{``The patient has been predicted to have pleural effusion (prediction 1) with maximum confidence (probability 1.00), based on consistent documentation of small bilateral pleural effusions across multiple recent radiology reports. The strongest evidence supporting this prediction comes from the most recent chest radiology examination [1], which explicitly states the presence of persistent bilateral pleural effusions. Additionally, another radiology report [3] notes "small bilateral pleural effusions," and a more detailed radiology note [4] similarly documents the presence of small bilateral pleural effusions in the context of postoperative cardiac surgery. These findings are consistent across different reports and imaging time points, reinforcing the model's high confidence in the prediction. No documents contradict the presence of pleural effusion.''}
\end{templatebox}

\begin{templatebox}[title=Example 2: Explanation for Mortality]
\footnotesize
\textit{``The model predicts low risk of mortality, likely because the patient's clinical trajectory demonstrates a modest but measurable improvement in pulmonary status [7] despite the presence of severe valvular heart disease and calcifications [2, 5]. The most recent chest radiograph shows a "mild decrease in pulmonary vascular congestion" and "slightly improved" pulmonary edema, indicating a positive response to ongoing treatment [7]. This improvement contrasts with earlier findings of "moderate pulmonary edema" and low lung volumes [10], suggesting a stabilizing or even improving course. While the echocardiogram documents critical aortic stenosis, severe mitral annular calcification, and moderate tricuspid regurgitation [2, 5], there is no evidence of acute decompensation, hemodynamic instability, or signs of tamponade from the small pericardial effusion [1, 3]. Additionally, imaging studies rule out acute intracranial hemorrhage, fracture, or major neurological compromise [8, 9], which would have significantly increased mortality risk. Although the patient has significant baseline cardiovascular disease, the model likely weighed the recent trend of pulmonary improvement and absence of acute catastrophic findings more heavily, leading to a low-risk classification.''}
\end{templatebox}

\section{Discussion}\label{sec3}

The xHAIM framework represents a significant advancement in medical AI, bridging generative and discriminative paradigms to address both performance and explainability challenges. We demonstrate that xHAIM improves predictive performance from 79.9\% to 90.3\% AUC across chest pathology tasks (Table~\ref{tab:auc_results}) while generating high-quality clinical explanations validated through rigorous evaluation (Tables~\ref{tab:eval_manual_detail} and \ref{tab:eval_llm_summary}). Our key contributions include: (a) a novel strategy that leverages generative AI to enhance—rather than replace—discriminative models, preserving their predictive strengths while adding interpretability; (b) significant performance gains through intelligent data curation that eliminates noise from embedding averaging; (c) clinically meaningful explanations that cite specific patient documents and medical knowledge, enhancing transparency; and (d) a practical framework that extends existing ML systems without requiring architectural overhauls. Furthermore, because predictions are well-calibrated by the discriminative models, the explanations are focused and accurately cite relevant patient documents, offering clinicians precise, actionable documentation and reducing administrative burden.

Beyond quantitative improvements, xHAIM addresses core technical challenges in healthcare AI through its integrated approach to data processing, prediction, and explanation. By prioritizing input quality via selective extraction and summarization, xHAIM improves not only AUC metrics but also the interpretability and trustworthiness of the system. The observed performance gains challenge the prevailing assumption that more data is always better—particularly in clinical contexts, where large volumes of unfiltered information may dilute the relevant signal. By emphasizing task-specific content, xHAIM achieves superior results, suggesting that intelligent curation may be more impactful than simply increasing model capacity or dataset size.

xHAIM’s explanation capabilities represent a substantial improvement over traditional methods. Unlike conventional post-hoc techniques that rely on opaque feature importance scores or attention weights, xHAIM provides explanations grounded in specific patient records and medical knowledge. This approach enhances transparency and fosters clinician trust, enabling effective human-AI collaboration. Moreover, xHAIM’s interactive explanations allow clinicians to inspect and validate the system’s reasoning, shifting AI from a static prediction engine to a dynamic decision-support partner.

The practical implementation demonstrates significant potential for enhancing clinical workflows without disruption. When clinicians access patient information, xHAIM automatically provides clinical summaries highlighting task-relevant findings, calibrated predictions with confidence indicators, and explanations with direct citations for rapid verification. This streamlined approach reduces chart review time while maintaining clinical authority over decisions. The framework's ability to operate with open-source models within hospital environments without external API dependencies further enhances its practical utility for healthcare institutions concerned with data privacy and security. The framework shows how AI can enhance rather than replace clinical expertise, providing efficiency gains without sacrificing the human elements essential to quality care.

While xHAIM demonstrates significant advances, several limitations merit acknowledgment. The quality of explanations depends on medical knowledge source availability and currency, particularly for rare conditions. Furthermore, the integration of LLMs at multiple stages introduces computational overhead, though rapid advances in open-source models are making such architectures increasingly viable. Finally, validation beyond MIMIC-IV across diverse populations and settings remains necessary. 

As healthcare continues generating increasingly complex data, approaches like xHAIM that efficiently process, filter, and explain this information will be essential for realizing AI's full potential in improving patient care. By bridging the gap between computational capability and clinical utility, xHAIM represents a crucial step toward more effective and trustworthy AI integration in healthcare settings.

\section{Methods}\label{sec4}

The xHAIM framework extends the original HAIM pipeline through a structured four-step process designed to enhance both predictive performance and interpretability, as shown in Figure~\ref{fig:pipeline}. First, it identifies and retrieves patient information relevant to the clinical task at hand. Second, it generates concise, task-specific summaries of this information. Third, it integrates these summaries across multiple data modalities and inputs them into a predictive model. Finally, it produces interpretable explanations that clarify the model’s predictions. In the sections that follow, we describe each of these steps in detail.

\subsection*{Modalities Preprocessing}
Building on the multimodal design of HAIM, our framework is general and can incorporate various data sources (modalities). The structured data is used as typically and the unstructured data is converted into natural language. Specifically, our framework supports:

\begin{itemize}
    \item \textbf{Tabular data}. Used directly as structured input.
    \item \textbf{Time series data}. Transformed into descriptive statistics that summarize the temporal nature of the data, such as minimum, maximum, mean, peaks etc. consistent with the original HAIM approach.
    \item \textbf{Clinical notes}. Includes radiology reports, past medical histories, discharge summaries, and other documentation.
    \item \textbf{Imaging data}. Images, together with their associated reports, are processed through the multimodal Qwen2.5-VL-72B\footnote{huggingface.co/Qwen/Qwen2.5-VL-72B-Instruct}\cite{bai2025qwen2} to generate comprehensive descriptions highlighting key findings.
\end{itemize}

This conversion of all unstructured modalities into natural language ensures compatibility with the generative LLM-based steps that follow.

\subsection*{Finding Relevant Chunks}
For each modality, we identify the segments most relevant to the prediction task. 
After splitting the document into manageable chunks, we define task-specific anchor sentences (containing e.g., "pneumonia", “consolidation,” and “infiltrate” for pneumonia) and score the semantic similarity between each chunk and the anchor using a hybrid metric:

\[
\text{Score}_{\text{hybrid}} = \alpha \cdot \text{BM25}_{\text{normalized}} + (1-\alpha) \cdot \text{SBERT}_{\text{sim}},
\]

\noindent
where $\alpha = 0.5$ balances keyword overlap with semantic similarity, BM25 is a TF–IDF–derived ranking function, and SBERT computes cosine similarity between sentence embeddings \citep{robertson2009probabilistic, reimers2019sentence}. We describe BM25 and SBERT in more detail in Appendix~\ref{subsecA1:retrieval}. From each modality, we retain the top-$k$ most relevant chunks to reduce noise and enhance the quality of subsequent summaries.

\subsection*{Generating Task-Specific Summaries}
The selected chunks are synthesized into coherent summaries using generative large language models (LLMs). We experimented with Llama-3.3-70B\footnote{huggingface.co/unsloth/Llama-3.3-70B-Instruct-bnb-4bit} \cite{grattafiori_llama_2024} and Qwen3-32B\footnote{huggingface.co/Qwen/Qwen3-32B} \cite{yang_qwen3_2025}, both open-source models that offer a favorable trade-off between performance and accessibility. Each of the unstructured modalities receives its own summary, that distills essential clinical content.

As an added benefit, this summarization step standardizes free-text notes, mitigating variability due to clinician writing styles or institutional documentation practices. As a result, the model becomes more robust to cross-hospital deployment.

\subsection*{Multimodal Integration} \label{subsec:multimodal_integration}
Next, we convert the generated summaries into embeddings suitable for downstream predictive modeling (e.g., XGBoost). For this, we fine-tune ClinicalBERT.

This is in contrast to HAIM’s use of frozen embeddings from raw, unfiltered text or image inputs; the original HAIM pipeline feeds raw notes and images through ClinicalBERT and a pre-trained Densenet121 Convolutional Neural Network (CNN) model, previously fine-tuned on the X-Ray CheXpert, respectively. Regarding the notes pipeline, the long, unfiltered text needs to be chunked into 512-token segments to fit in the ClinicalBERT model, which results in two main problems; multiple embeddings need to be averaged, resulting in potential introduction of noise, and also performing finetuning is not straightforward, as not all parts of the notes contain meaningful information for the corresponding label. As a result, the extracted frozen embeddings may fail to capture clinically relevant information in a form that is most useful for the downstream prediction tasks.

By contrast, our summaries are optimized to fit within ClinicalBERT's input length, allowing for effective fine-tuning. We train separate ClinicalBERT models for each summary type, discarding classification heads post-training to retain only the embeddings for integration.

There exist different integration strategies that can be employed, with the most straightforward being: (1) combining all summaries into a single document to produce one embedding, and (2) generating separate embeddings per modality and concatenating them. We adopt the latter, which mirrors HAIM’s original architecture and facilitates modality-specific performance analysis.

Thus, the final feature representation is:
\[
\mathbf{X} = [\mathbf{x}_{\text{notes\_summary}}, \mathbf{x}_{\text{cxr\_summary}}, \mathbf{x}_{\text{tabular}}, \mathbf{x}_{\text{time\_series}}],
\]
where $\mathbf{x}_{\text{notes\_summary}}$ and $\mathbf{x}_{\text{cxr\_summary}}$ are 768-dimensional ClinicalBERT embeddings corresponding to the [CLS] token, and $\mathbf{x}_{\text{tabular}}$ and $\mathbf{x}_{\text{time\_series}}$ are statistical feature vectors. 

\subsection*{Explanation Generation}\label{subsec:explanation}

The final step of xHAIM generates interpretable explanations by combining three inputs: patient summaries, model predictions, and relevant medical knowledge. This approach addresses LLM limitations in clinical tasks, such as difficulty handling long records and susceptibility to hallucinations.
By grounding explanation generation in curated summaries, calibrated predictive outputs, and authoritative domain knowledge, we ensure that the explanations are accurate, transparent, and clinically meaningful.

The generated explanations reference original patient content and relevant clinical criteria, positioning the model as a decision-support tool rather than an opaque black box. Clinicians can inspect citations, understand the reasoning behind predictions, and retain decision-making authority, aiming to increase trust and adoption.

In sum, our framework systematically improves upon HAIM by integrating generative LLMs for targeted summarization, enabling fine-tuned multimodal embeddings, and producing grounded, interpretable explanations.

\subsection*{Code availability}

The code for the xHAIM framework will be made available upon publication at \href{https://github.com/PericlesPet/xHAIM}{https://github.com/PericlesPet/xHAIM}. This includes all implementations for document retrieval, summary generation, multimodal integration, training and finetuning, explanation generation, and automatic explanation evaluation.

\backmatter

\bmhead{Supplementary information}

Supplementary materials include additional experimental results, ablation studies, and example outputs from the xHAIM system.

\bmhead{Acknowledgements}

The authors would like to acknowledge the MIMIC project for providing the data used in this study. Special thanks to the authors of the original HAIM framework for their contributions, and for making their code publicly available.

\section*{Declarations}

\begin{itemize}
\item Funding: This work was conducted by the authors while they were affiliated with the Massachusetts Institute of Technology.
\item Conflict of interest/Competing interests: The authors declare no competing interests.
\item Ethics approval and consent to participate: Not applicable.
\item Consent for publication: Not applicable.
\item Data availability: The HAIM-MIMIC-MM dataset used in this study is publicly available through PhysioNet, subject to completing appropriate training and data use agreements.
\item Materials availability: Not applicable.
\item Code availability: The code for implementing the xHAIM framework will be made available upon publication at \href{https://github.com/PericlesPet/xHAIM}{https://github.com/PericlesPet/xHAIM}.
\item Author contribution: P.P. and G.M. developed the methodology and experimental framework; P.P, G.M and V.S. conducted the experiments; P.P and V.S. performed the explanations annotations; V.S. prepared the data; D.B. directed the overall research; all authors contributed to writing and editing the manuscript.
\end{itemize}

\bigskip

\bibliography{sn-bibliography}


\end{document}